\documentclass{article}

\usepackage[final]{neurips_2023}

\usepackage[utf8]{inputenc} 
\usepackage[T1]{fontenc}    
\usepackage{hyperref}       
\usepackage{url}            
\usepackage{booktabs}       
\usepackage{amsfonts}       
\usepackage{nicefrac}       
\usepackage{microtype}      
\usepackage{xcolor}         
\usepackage{graphicx}
\usepackage{nicematrix,enumitem}

\title{Cancer-Net PCa-Data: An Open-Source Benchmark Dataset for Prostate Cancer Clinical Decision Support using Synthetic Correlated Diffusion Imaging Data}

\author{%
  Hayden Gunraj\\
  Department of Systems Design Engineering\\
  University of Waterloo \\
  \texttt{hayden.gunraj@uwaterloo.ca} \\
  \And
  Chi-en A. Tai \\
  Department of Systems Design Engineering\\
  University of Waterloo\\
  Waterloo, ON\\
  \texttt{amy.tai@uwaterloo.ca} \\
  \And
  Alexander Wong \\
  Department of Systems Design Engineering\\
  University of Waterloo \\
  \texttt{alexander.wong@uwaterloo.ca} \\
}

\begin{document}

\maketitle

\begin{abstract}
The recent introduction of synthetic correlated diffusion (CDI\textsuperscript{s}) imaging has demonstrated significant potential in the realm of clinical decision support for prostate cancer (PCa). CDI\textsuperscript{s} is a new form of magnetic resonance imaging (MRI) designed to characterize tissue characteristics through the joint correlation of diffusion signal attenuation across different Brownian motion sensitivities. Despite the performance improvement, the CDI\textsuperscript{s} data for PCa has not been previously made publicly available. In our commitment to advance research efforts for PCa, we introduce Cancer-Net PCa-Data, an open-source benchmark dataset of volumetric CDI\textsuperscript{s} imaging data of PCa patients. Cancer-Net PCa-Data consists of CDI\textsuperscript{s} volumetric images from a patient cohort of 200 patient cases, along with full annotations (gland masks, tumor masks, and PCa diagnosis for each tumor). We also analyze the demographic and label region diversity of Cancer-Net PCa-Data for potential biases. Cancer-Net PCa-Data is the first-ever public dataset of CDI\textsuperscript{s} imaging data for PCa, and is a part of the global open-source initiative dedicated to advancement in machine learning and imaging research to aid clinicians in the global fight against cancer.
\end{abstract}

\section{Introduction}
The recent introduction of synthetic correlated diffusion (CDI\textsuperscript{s}) imaging has demonstrated significant potential in the realm of clinical decision support for prostate cancer (PCa)~\cite{wong2022synthetic}. CDI\textsuperscript{s} is a new form of magnetic resonance imaging (MRI) designed to characterize tissue characteristics through the joint correlation of diffusion signal attenuation across different Brownian motion sensitivities.  This is achieved by capturing diffusion signal acquisitions using different gradient pulse strengths and timings, and mixing both native and synthetic diffusion signal acquisitions together in a calibrated fashion.  As such, CDI\textsuperscript{s} allows for the quantification of water molecule distribution with respect to their degree of Brownian motion within tissue.  Compared to current standard MRI techniques such as T2-weighted imaging (T2w), diffusion-weighted imaging (DWI), and dynamic contrast-enhanced imaging (DCE), statistical analyses revealed that CDI\textsuperscript{s} achieves stronger delineation of clinically significant cancerous tissue and is a stronger indicator of PCa presence~\cite{wong2022synthetic}. 

 \begin{figure}[t]
  \centering
   \includegraphics[width=\linewidth]{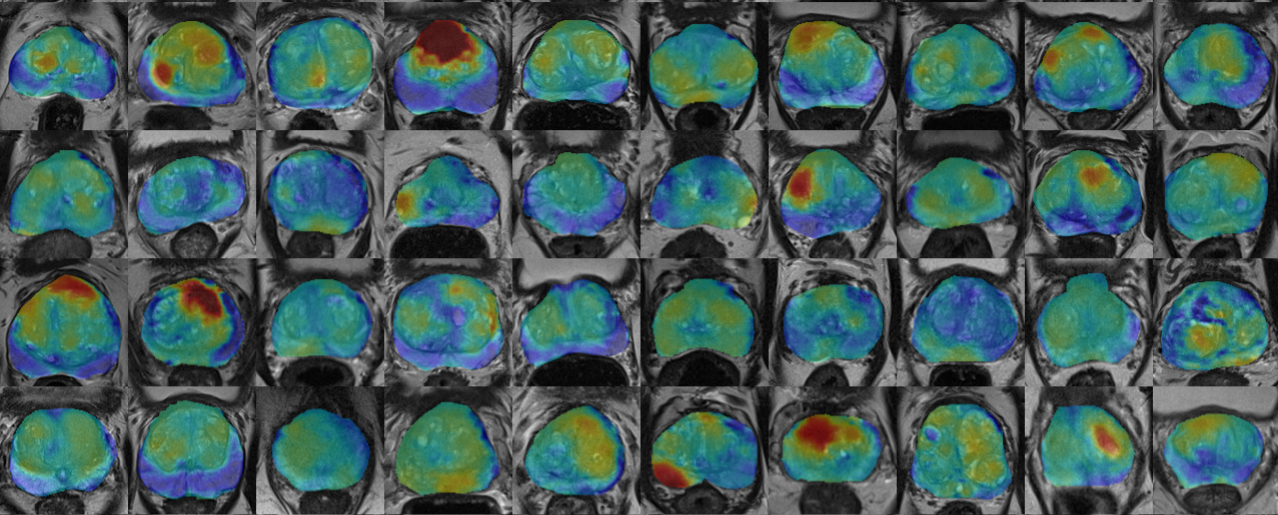}
   \caption{Example CDI\textsuperscript{s} data from patient cohort in Cancer-Net PCa-Data.}
   \label{fig:cancernetpcadata}
\end{figure}

\begin{figure}[b]
  \centering
   \includegraphics[width=\linewidth]{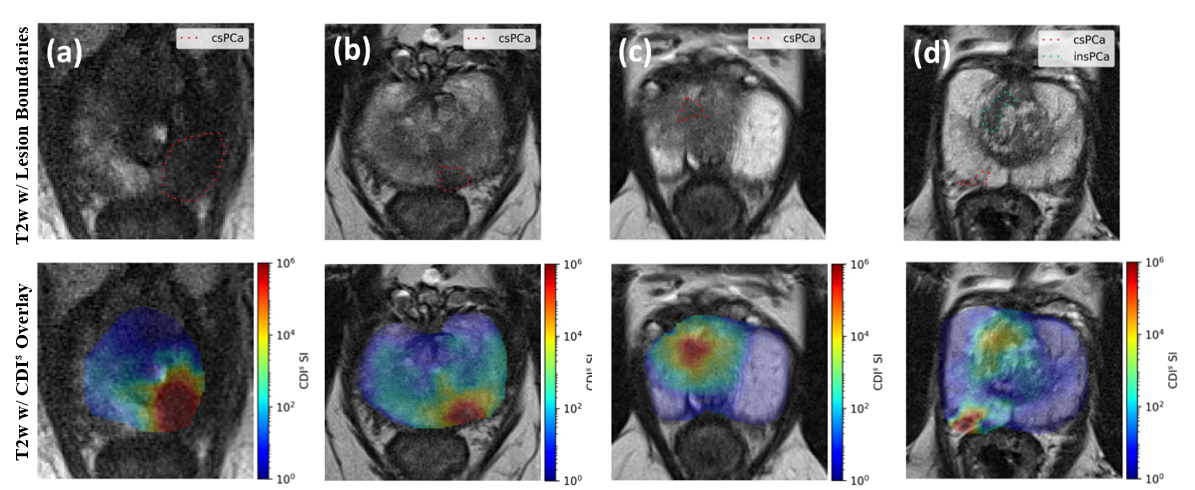}
   \caption{T2w images with overlays of annotated tumour boundaries in dotted lines and CDI\textsuperscript{s} in color overlay for six patient cases. Cancer-Net PCa-Data contains prostate masks, tumor masks, and tumor annotations (Gleason score). (a, b) Two patients with csPCa in the peripheral zone. (c) A patient with csPCa in the transition zone. (d) A patient with csPCa in the peripheral zone and insPCa in the transition zone.}
   \label{fig:grade-ex}
\end{figure}

Despite the potential impact of CDI\textsuperscript{s} for PCa, the CDI\textsuperscript{s} patient data for PCa has not been previously made publicly available. In our commitment to advance research efforts for PCa, we introduce Cancer-Net PCa-Data, an open-source benchmark dataset of volumetric CDI\textsuperscript{s} imaging data of PCa patients with the associated label regions of healthy, clinically significant, and clinically insignificant prostate cancer. We also analyze the demographic and label region diversity of the Cancer-Net PCa-Data dataset for potential biases. Cancer-Net PCa-Data is the first-ever public dataset of CDI\textsuperscript{s} imaging data for PCa, and is a part of the global open-source initiative dedicated to advancement in machine learning and imaging research to aid clinicians in the global fight against cancer and has been made publicly available at \url{https://www.kaggle.com/datasets/hgunraj/cancer-net-pca-data}.

\section{Methodology}
The Cancer-Net PCa-Data dataset compromises of CDI\textsuperscript{s} imaging data computed from a patient cohort of 200 patient cases acquired at Radboud University Medical Centre (Radboudumc) in the Prostate MRI Reference Center in Nijmegen, The Netherlands and made available as part of the SPIE-AAPM-NCI PROSTATEx Challenges~\cite{Litjens2017, Litjens2014,Clark2013,Cuocolo2021} (see Figure~\ref{fig:cancernetpcadata} for example CDI\textsuperscript{s} data from the patient cohort). Masks derived from the PROSTATEx\_masks repository are also provided and indicate which label regions of healthy prostate tissue, clinically significant prostate cancer (csPCa), and clinically insignificant prostate cancer (insPCa)~\cite{Litjens2017, Litjens2014,Clark2013,Cuocolo2021}. Tumours with a Prostate Imaging-Reporting and Data System (PI-RADS) score of 1 or 2 were considered clinically insignificant and not biopsied. Annotations for prostate cancer, whole gland, transition zone, and peripheral zone were performed by two radiology residents and two experienced board-certified radiologists working in pairs at the University of Naples Federico II, Naples, Italy~\cite{cuocolo2021quality}.

195 patients (97.5\%) were imaged using a Siemens MAGNETOM Skyra 3.0T machine and 5 patient acquisitions (2.5\%) were obtained from a Siemens MAGNETOM Trio 3.0T machine~\cite{Litjens2014}. An expert radiologist who over 20 years of experience interpreting prostate MRI reviewed or supervised the acquisition of these images~\cite{Litjens2014}. Axial DWI acquisitions used a single-shot echo-planar sequence with a TR range of 2500-3300 ms (median of 2700 ms), TE range of 63-81 ms (median of 63 ms), and in-plane resolution of 2 mm with the slice thickness range of 3-4.5 mm (median = 3 mm), at three b-values (50 s/mm\textsuperscript{2}, 400 s/mm\textsuperscript{2}, 800 s/mm\textsuperscript{s}) with the display field of view range 16.8x25.6 cm\textsuperscript{2} to 24.0x25.6 cm\textsuperscript{2} (median of 16.8x25.6 cm\textsuperscript{2}~\cite{Litjens2014}. CDI\textsuperscript{s} was computed from native and synthetic DWI signal acquisitions across 9 b-values (0, 50, 400, 800, 1000, 2000, 3000, 4000, 5000).  Examples of different patient cases and their corresponding CDI\textsuperscript{s} data, tumor masks, and PCa diagnosis are shown in Figure~\ref{fig:grade-ex}.  

\section{Results and Discussion}
The demographics of the Cancer-Net PCa-Data dataset is shown in Table~\ref{tab:demographics}. The patients range from 37 to 78 years with a median age of 64 years. As seen, the 60-69 age range dominates the data with over half of the patient cohort falling into this age range (56\%), illustrating a potential bias towards this age range. In addition, there are very few patients younger than 50 in the dataset, illustrating that this population is very underrepresented in the dataset. 

\begin{table}[t!]
  \caption{Summary of age demographic in the Cancer-Net PCa-Data dataset.}
  \centering
  \NiceMatrixOptions{notes/para}
    \begin{NiceTabular}{l r r}
        \toprule
        \RowStyle{\bfseries}
    Age & Number of Patients & Percentage \\ \midrule
    30-39 & 3 & 1.5\% \\
    40-49 & 5 & 2.5\% \\
    50-59 & 45 & 22.5\% \\
    60-69 & 112 & 56\% \\
    70-79 & 35 & 17.5\% \\
    \bottomrule
    \end{NiceTabular}
  \label{tab:demographics}
\end{table}

The clinical significance is shown in Table~\ref{tab:clinicalsignificance} and is presented at the tumour level (rather than the patient level). There is an uneven distribution between clinically significant and clinically insignificant tumours in the dataset, with almost three times more clinically insignificant tumours than clinically significant tumours. Noting this class imbalance, is is recommended to use data sampling, re-balance the classes, and/or implement a balanced loss function to mitigate model training issues. When evaluating systems developed on this dataset, balanced metrics such as per-class precision and recall should also be implemented to account for the class imbalance. 

\begin{table}[t!]
  \caption{Summary of clinical significance variable in the patient cohort (tumour level).}
  \centering
  \NiceMatrixOptions{notes/para}
    \begin{NiceTabular}{l r r}
        \toprule
        \RowStyle{\bfseries}
    Clinical Significance & Number of Tumours & Percentage \\ \midrule
    csPCa & 76 & 25.4\% \\
    insPCa & 223 & 74.6\% \\
    \bottomrule
    \end{NiceTabular}
  \label{tab:clinicalsignificance}
\end{table}

\begin{figure}[!ht]
  \centering
   \includegraphics[width=\linewidth]{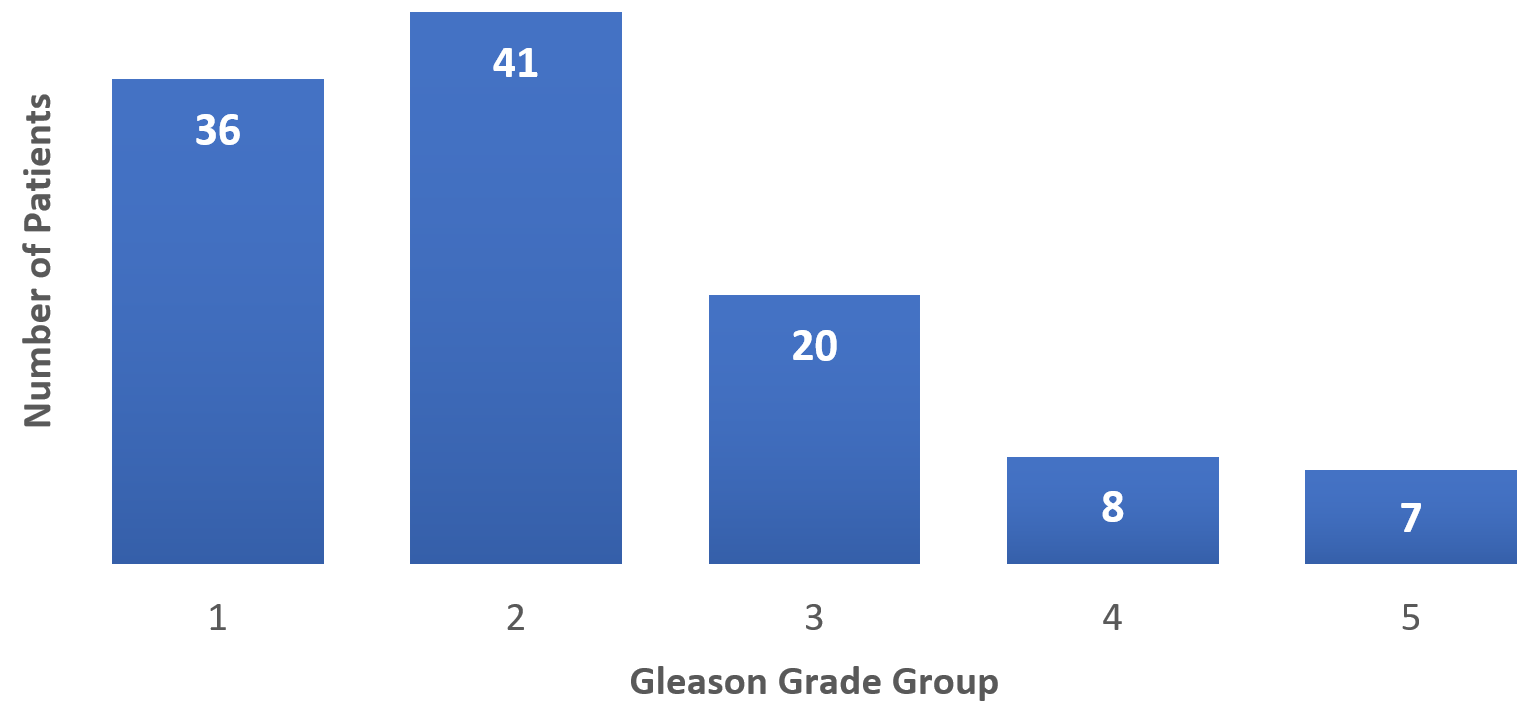}
   \caption{Distribution of the Gleason Grade for tumours that were biopsied in Cancer-Net PCa-Data.}
   \label{fig:gleason-grade-dist}
\end{figure}

Unfortunately the Gleason Grade Group for most of the tumours (187 tumours out of 299 tumours or 62.5\%) is not available (no biopsy information). The information may have not been recorded or in cases of clinically insignificant prostate cancers, these tumours were not biopsied. However, the distribution for the other tumours is shown in Figure~\ref{fig:gleason-grade-dist}. As seen, there is more tumour samples for Gleason Grade Group 1 and 2 compared to Group 4 and 5. However, the numerous entries with no biopsy information makes it difficult to make any significant claim about this variable. Regardless, this uneven distribution is still important to note when using this dataset for model development, especially if the task is to classify the Gleason Grade of prostate cancer tumours.

\section{Conclusion}
In this work, we introduce Cancer-Net PCa-Data, an open-source benchmark dataset of volumetric CDI\textsuperscript{s} imaging data of PCa patients. We analyze the demographic and label region variables that are present in the dataset and highlight potential biases. Specifically, we notice more data for the 60 - 69 years age group and more clinically insignificant tumour data compared to clinically significant tumour data. Given the class imbalances, we recommend leveraging various algorithms and strategies such as data sampling, re-balancing of the classes, and balanced loss functions and evaluating systems developed on this dataset using balanced metrics such as per-class precision and recall.

\section{Potential Negative Societal Impact}
Potential negative societal impacts include misusing the data and over-reliance on models trained on this dataset. Though the motivation of open-sourcing this benchmark dataset is to support research advancements in this field, it is possible for others to misuse the data by building algorithms using the data to adjust insurance premiums based on forecasted medical expenses for individual patients. On the other hand, using the data to train models for clinical use can also have a detrimental impact if there is over-reliance on the model's results and it is not properly validated nor continually retrained. Subsequently, we encourage the validation of any models trained using this dataset to be validated on real-world clinical data and to be used with expert oversight. 

\begin{ack}
The authors thank the Natural Sciences and Engineering Research Council of Canada and the Canada Research Chairs Program.
\end{ack}

{
\small

\bibliography{neurips_2023}

\begin{thebibliography}{1}

\bibitem{wong2022synthetic}
Alexander Wong, Hayden Gunraj, Vignesh Sivan, and Masoom~A Haider.
\newblock Synthetic correlated diffusion imaging hyperintensity delineates clinically significant prostate cancer.
\newblock {\em Scientific Reports}, 12(1):3376, 2022.

\bibitem{Litjens2017}
Geert Litjens, Oscar Debats, Jelle Barentsz, Nico Karssemeijer, and Henkjan Huisman.
\newblock Prostatex challenge data [data set], 2017.

\bibitem{Litjens2014}
Geert Litjens, Oscar Debats, Jelle Barentsz, Nico Karssemeijer, and Henkjan Huisman.
\newblock Computer-aided detection of prostate cancer in mri.
\newblock {\em IEEE Transactions on Medical Imaging}, 33(5):1083--1092, 2014.

\bibitem{Clark2013}
Kenneth Clark, Bruce Vendt, Kirk Smith, John Freymann, Justin Kirby, Paul Koppel, Stephen Moore, Stanley Phillips, David Maffitt, Michael Pringle, Lawrence Tarbox, and Fred Prior.
\newblock The cancer imaging archive (tcia): Maintaining and operating a public information repository.
\newblock {\em Journal of Digital Imaging}, 26(6):1045--1057, 2013.

\bibitem{Cuocolo2021}
Renato Cuocolo, Arnaldo Stanzione, Anna Castaldo, Davide~Raffaele {De Lucia}, and Massimo Imbriaco.
\newblock Quality control and whole-gland, zonal and lesion annotations for the prostatex challenge public dataset.
\newblock {\em European Journal of Radiology}, 138:109647, 2021.

\bibitem{cuocolo2021quality}
Renato Cuocolo, Arnaldo Stanzione, Anna Castaldo, Davide~Raffaele De~Lucia, and Massimo Imbriaco.
\newblock Quality control and whole-gland, zonal and lesion annotations for the prostatex challenge public dataset.
\newblock {\em European Journal of Radiology}, 138:109647, 2021.

\end{thebibliography}
}

\end{document}